\def\year{2017}\relax
\documentclass[letterpaper]{article}
\usepackage{aaai17}
\usepackage{times}
\usepackage{helvet}
\usepackage{courier}
\usepackage{multirow}
\usepackage{fancyvrb}
\usepackage{graphicx}
\usepackage{float}
\usepackage{subcaption}
\usepackage{amsmath, amssymb}
\usepackage{wrapfig}
\begin{document}
\title{Attention Correctness in Neural Image Captioning}
\author{Chenxi Liu$^1$ \quad Junhua Mao$^2$ \quad Fei Sha$^{2, 3}$ \quad Alan Yuille$^{1, 2}$ \\
Johns Hopkins University$^1$ \\
University of California, Los Angeles$^2$ \\
University of Southern California$^3$}
\maketitle

\begin{abstract}
Attention mechanisms have recently been introduced in deep learning for various tasks in natural language processing and computer vision. But despite their popularity, the ``correctness'' of the implicitly-learned attention maps has only been assessed qualitatively by visualization of several examples. In this paper we focus on evaluating and improving the correctness of attention in neural image captioning models. Specifically, we propose a quantitative evaluation metric for the consistency between the generated attention maps and human annotations, using recently released datasets with alignment between regions in images and entities in captions. We then propose novel models with different levels of explicit supervision for learning attention maps during training. The supervision can be strong when alignment between regions and caption entities are available, or weak when only object segments and categories are provided. We show on the popular Flickr30k and COCO datasets that introducing supervision of attention maps during training solidly improves both attention correctness and caption quality, showing the promise of making machine perception more human-like. 
\end{abstract}

\section{Introduction}

Recently, attention based deep models have been proved effective at handling a variety of AI problems such as machine translation \cite{bahdanau2014neural}, object detection \cite{mnih2014recurrent,ba2014multiple}, visual question answering \cite{xu2015ask,chen2015abc}, and image captioning \cite{xu2015show}. % Action Recognition? 
Inspired by human attention mechanisms, these deep models learn dynamic weightings of the input vectors, which allow for more flexibility and expressive power.

In this work we focus on attention models for image captioning. 
The state-of-the-art image captioning models \cite{kiros2014unifying,mao2014deep,karpathy2015deep,donahue2015long,vinyals2015show} adopt Convolutional Neural Networks (CNNs) to extract image features and Recurrent Neural Networks (RNNs) to decode these features into a sentence description.
Within this encoder-decoder framework \cite{cho2014learning}, the models proposed by \cite{xu2015show} apply an attention mechanism, i.e. attending to different areas of the image when generating words one by one.

Although impressive visualization results of the attention maps for image captioning are shown in \cite{xu2015show}, the authors do not provide \textit{quantitative evaluations} of the attention maps generated by their models.
Since deep network attention can be viewed as a form of alignment from language space to image space, we argue that these attention maps in fact carry important information in understanding (and potentially improving) deep networks.
Therefore in this paper, we study the following two questions:
\begin{itemize}
\item How often and to what extent are the attention maps consistent with human perception/annotation?
\item Will more human-like attention maps result in better captioning performance?
\end{itemize}

\begin{figure}[t]
\centering
\includegraphics[width=0.9\linewidth]{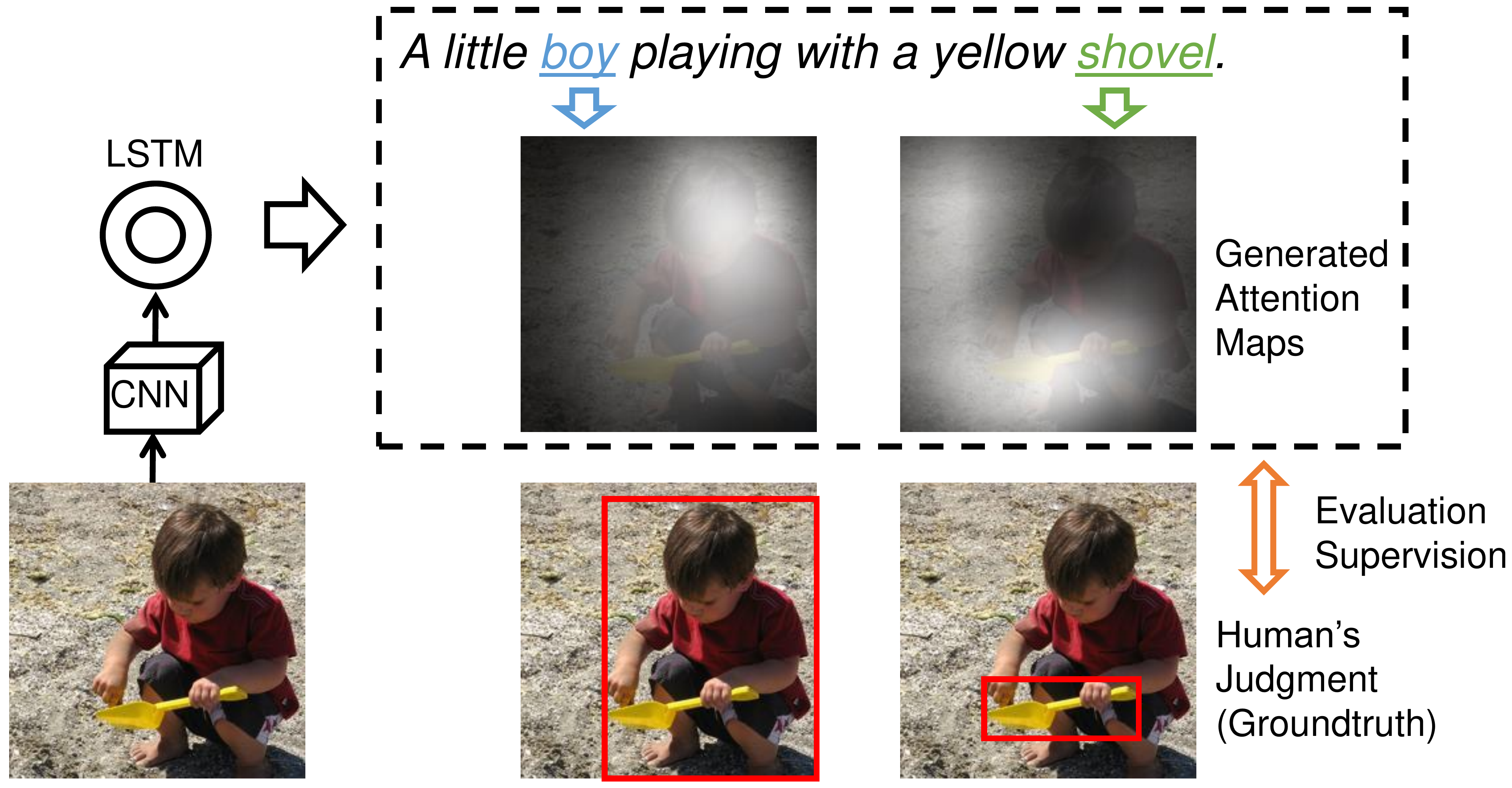}
\caption{Image captioning models \protect\cite{xu2015show} can attend to different areas of the image when generating the words. However, these generated attention maps may not correspond to the region that the words or phrases describe in the image (e.g. ``shovel''). We evaluate such phenomenon quantitatively by defining attention correctness, and alleviate this inconsistency by introducing explicit supervision. In addition, we show positive correlation between attention correctness and caption quality.}
\label{fig:overview}
\end{figure}

Towards these goals, we propose a novel quantitative metric to evaluate the ``correctness'' of attention maps.
We define ``correctness'' as the consistency between the attention maps generated by the model and the corresponding region that the words/phrases describe in the image.
More specifically, we use the alignment annotations between image regions and noun phrase caption entities provided in the Flickr30k Entities dataset \cite{plummer2015flickr30k} as our ground truth maps.
Using this metric, we show that the attention model of \cite{xu2015show} performs better than the uniform attention baseline, but still has room for improvement in terms of attention consistency with human annotations.

Based on this observation, we propose a model with explicit supervision of the attention maps.
The model can be used not only when detailed ground truth attention maps are given (e.g. the Flickr30k Entities dataset \cite{plummer2015flickr30k}) but also when only the semantic labelings of image regions (which is a much cheaper type of annotations) are available (e.g. MS COCO dataset \cite{lin2014microsoft}).
Our experiments show that in both scenarios, our models perform consistently and significantly better than the implicit attention counterpart in terms of both attention maps accuracy and the quality of the final generated captions.
To the best of our knowledge, this is the first work that quantitatively measures the quality of visual attention in deep models and shows significant improvement by adding supervision to the attention module.

\section{Related Work}

\textbf{Image Captioning Models}
There has been growing interest in the field of image captioning, with lots of work demonstrating impressive results \cite{kiros2014unifying,xu2015show,mao2014deep,vinyals2015show,donahue2015long,fang2015captions,karpathy2015deep,chen2014learning}.
However, it is uncertain to what extent the captioning models truly understand and recognize the objects in the image while generating the captions.
\cite{xu2015show} proposed an attention model and qualitatively showed that the model can attend to specific regions of the image by visualizing the attention maps of a few images.
Our work takes a step further by quantitatively measuring the quality of the attention maps.
The role of the attention maps also relates to referring expressions \cite{mao2015generation,hu2015natural}, where the goal is predicting the part of the image that is relevant to the expression.

\noindent
\textbf{Deep Attention Models}
% Attention mechanisms are important properties of human visual systems \cite{rensink2000dynamic,corbetta2002control}. 
% Since deep neural networks are inspired by the structure of neurons in human brains, exploring the use of attention in these artificial models seems natural and promising.
In machine translation, \cite{bahdanau2014neural} introduced an extra softmax layer in the RNN/LSTM structure that generates weights of the individual words of the sentence to be translated. 
The quality of the attention/alignment was qualitatively visualized in \cite{bahdanau2014neural} and quantitatively evaluated in \cite{luong2015effective} using the alignment error rate. 
In image captioning, \cite{xu2015show} used convolutional image features with spatial information as input, allowing attention on 2D space.
\cite{you2016image} targeted attention on a set of concepts extracted from the image to generate image captions.
In visual question answering, \cite{chen2015abc,xu2015ask,shih2016wtl,zhu2015visual7w} proposed several models which attend to image regions or questions when generating an answer.
But none of these models quantitatively evaluates the quality of the attention maps or imposes supervision on the attention.
Concurrently, \cite{das2016human} analyzed the consistency between human and deep network attention in visual question answering.
Our goal differs in that we are interested in how attention changes with the progression of the description.
% Eye tracking cannot be applied trivially in this case, because description generation lags behind eye movements \cite{griffin2000eyes}.

\noindent
\textbf{Image Description Datasets} 
For image captioning, Flickr8k \cite{hodosh2013framing}, Flickr30k \cite{young2014image}, and MS COCO \cite{lin2014microsoft} are the most commonly used benchmark datasets. 
% Each image in these datasets has 5 accompanying captions.
% The original annotations of these datasets do not have alignment between the image regions and the entities (e.g. noun phrases) in the captions.
\cite{plummer2015flickr30k} developed the original caption annotations in Flickr30k by providing the region to phrase correspondences.
Specifically, annotators were first asked to identify the noun phrases in the captions, and then mark the corresponding regions with bounding boxes.
In this work we use this dataset as ground truth to evaluate the quality of the generated attention maps, as well as to train our strongly supervised attention model. 
Our model can also utilize the instance segmentation annotations in MS COCO to train our weakly supervised version. 

\section{Deep Attention Models for Image Captioning}
\label{sec:model}

In this section, we first discuss the attention model that learns the attention weights implicitly \cite{xu2015show}, and then introduce our explicit supervised attention model.

\subsection{Implicit Attention Model}
\label{sec:implicit_model}

The implicit attention model \cite{xu2015show} consists of three parts: the encoder which encodes the visual information (i.e. a visual feature extractor), the decoder which decodes the information into words, and the attention module which performs spatial attention.

The visual feature extractor produces $L$ vectors that correspond to different spatial locations of the image:
$a = \{\mathbf{a}_1, \hdots, \mathbf{a}_L\}, \ \mathbf{a}_i \in \mathbb{R}^D$.
Given the visual features, the goal of the decoder is to generate a caption $y$ of length $C$: $y = \{y_1, \hdots, y_C\}$. We use $\mathbf{y}_t \in \mathbb{R}^K$ to represent the one-hot encoding of $y_t$, where $K$ is the dictionary size.

In \cite{xu2015show}, an LSTM network \cite{hochreiter1997long} is used as the decoder:
\begin{align}
&\mathbf{i}_t = \sigma(W_i E \mathbf{y}_{t-1} + U_i \mathbf{h}_{t-1} + Z_i \mathbf{z}_t + \mathbf{b}_i) \\
&\mathbf{f}_t = \sigma(W_f E \mathbf{y}_{t-1} + U_f \mathbf{h}_{t-1} + Z_f \mathbf{z}_t + \mathbf{b}_f) \\
&\mathbf{c}_t = \mathbf{f}_t \mathbf{c}_{t-1} + \mathbf{i}_t \text{tanh}(W_c E \mathbf{y}_{t-1} + U_c \mathbf{h}_{t-1} + Z_c \mathbf{z}_t + \mathbf{b}_c) \\
&\mathbf{o}_t = \sigma(W_o E \mathbf{y}_{t-1} + U_o \mathbf{h}_{t-1} + Z_o \mathbf{z}_t + \mathbf{b}_o) \\
&\mathbf{h}_t = \mathbf{o}_t \text{tanh}(\mathbf{c}_t)
\end{align}
where $\mathbf{i}_t, \mathbf{f}_t, \mathbf{c}_t, \mathbf{o}_t, \mathbf{h}_t$ are input gate, forget gate, memory, output gate, and hidden state of the LSTM respectively. 
$W, U, Z, \mathbf{b}$ are weight matrices and biases.
% $\hat{\mathbf{z}}_t$ is a context vector derived from visual features $a$ as described below.
$E \in \mathbb{R}^{m \times K}$ is an embedding matrix, and $\sigma$ is the sigmoid function. 
The context vector $\mathbf{z}_t = \sum_{i=1}^L \alpha_{ti} \mathbf{a}_i$ is a dynamic vector that represents the relevant part of image feature at time step $t$, where $\alpha_{ti}$ is a scalar weighting of visual vector $\mathbf{a}_i$ at time step $t$, defined as follows:
\begin{equation}
\alpha_{ti} = \frac{\exp(e_{ti})}{\sum_{k=1}^L \exp(e_{tk})} \quad \quad e_{ti} = f_{attn}(\mathbf{a_i}, \mathbf{h}_{t-1}) 
\end{equation}
$f_{attn}(\mathbf{a_i}, \mathbf{h}_{t-1})$ is a function that determines the amount of attention allocated to image feature $\mathbf{a_i}$, conditioned on the LSTM hidden state $\mathbf{h}_{t-1}$.
In \cite{xu2015show}, this function is implemented as a multilayer perceptron. Note that by construction $\sum_{i=1}^L \alpha_{ti} = 1$. 

The output word probability is determined by the image $\mathbf{z}_t$, the previous word $y_{t-1}$, and the hidden state $\mathbf{h}_t$:
\begin{equation}
p(y_t| a, y_{t-1}) \propto \exp( G_o (E \mathbf{y}_{t-1} + G_h \mathbf{h}_t + G_z \mathbf{z}_t))
\end{equation}
where $G$ are learned parameters. The loss function, ignoring the regularization terms, is the negative log probability of the ground truth words $w = \{w_1, \hdots, w_C\}$:
\begin{equation} \label{eqn:loss-cap}
L_{t, cap} = -\log p(w_t | a, y_{t-1})
\end{equation}

\subsection{Supervised Attention Model}

In this work we are interested in the attention map generated by the model $\pmb{\alpha}_t = \{\alpha_{ti}\}_{i = 1, \hdots, L}$.
One limitation of the model in \cite{xu2015show} is that even if we have some prior knowledge about the attention map, it will not be able to take advantage of this information to learn a better attention function $f_{attn}(\mathbf{a_i}, \mathbf{h}_{t-1})$.
We tackle this problem by introducing explicit supervision.

Concretely, we first consider the case when the ground truth attention map $\pmb{\beta}_t = \{\beta_{ti}\}_{i = 1, \hdots, L}$ is provided for the ground truth word $w_t$, with $\sum_{i=1}^L \beta_{ti} = 1$. 
Since $\sum_{i=1}^L \beta_{ti} = \sum_{i=1}^L \alpha_{ti} = 1$, they can be considered as two probability distributions of attention and it is natural to use the cross entropy loss.
For the words that do not have an alignment with an image region (e.g. ``a'', ``is''), we simply set $L_{t, attn}$ to be 0:
\begin{equation}
L_{t, attn} = 
\begin{cases} 
-\sum_{i=1}^L \beta_{ti} \log \alpha_{ti} & \text{if } \pmb{\beta}_t \text{ exists for } w_t \\
0 & \text{otherwise}
\end{cases}
\end{equation}
The total loss is the weighted sum of the two loss terms: $L = \sum_{t=1}^C L_{t, cap} + \lambda \sum_{t=1}^C L_{t, attn}$.

We then discuss two ways of constructing the ground truth attention map $\pmb{\beta}_t$, depending on the types of annotations. % available. 

\subsubsection{Strong Supervision with Alignment Annotation}
\label{sec:model-strong}

In the simplest case, we have direct annotation that links the ground truth word $w_t$ to a region $R_t$ (in the form of bounding boxes or segmentation masks) in the image (e.g. Flickr30k Entities).
We encourage the model to ``attend to'' $R_t$ by constructing $\pmb{\hat{\beta}}_t = \{\hat{\beta}_{t\hat{i}}\}_{\hat{i}=1, .., \hat{L}}$ where:
\begin{equation}
\hat{\beta}_{t\hat{i}} = \begin{cases}
1 & \hat{i} \in R_t \\
0 & \text{otherwise}
\end{cases}
\end{equation}
Note that the resolution of the region $R$ (e.g. $224 \times 224$) and the attention map $\pmb{\alpha}, \pmb{\beta}$ (e.g. $14 \times 14$) may be different, so $\hat{L}$ could be different from $L$. 
Therefore we need to resize $\pmb{\hat{\beta}}_t$ to the same resolution as $\pmb{\alpha}_t$ and normalize it to get $\pmb{\beta}_t$. 

\subsubsection{Weak Supervision with Semantic Labeling}
\label{sec:model-weak}

Ground truth alignment is expensive to collect and annotate.
A much more general and cheaper annotation is to use bounding boxes or segmentation masks with object class labels (e.g. MS COCO).
In this case, we are provided with a set of regions $R_j$ in the image with associated object classes $c_j$, $j=1, \hdots, M$ where $M$ is the number of object bounding boxes or segmentation masks in the image. 
Although not ideal, these annotations contain important information to guide the attention of the model.
For instance, for the caption ``a boy is playing with a dog'', the model should attend to the region of a person when generating the word ``boy'', and attend to the region of a dog when generating the word ``dog''.
This suggests that we can approximate image-to-language (region $\rightarrow$ word) consistency by language-to-language (object class $\rightarrow$ word) similarity.

Following this intuition, we set the likelihood that a word $w_t$ and a region $R_j$ are aligned by the similarity of $w_t$ and $c_j$ in the word embedding space:
\begin{equation} \label{eqn:sim}
\hat{\beta}_{t\hat{i}} = \begin{cases}
\text{sim}(\tilde{E}(w_t), \tilde{E}(c_j)) & \hat{i} \in R_j \\
0 & \text{otherwise}
\end{cases}
\end{equation}
where $\tilde{E}(w_t)$ and $\tilde{E}(c_j)$ denote the embeddings of the word $w_t$ and $c_j$ respectively.
$\tilde{E}$ can be the embedding $E$ learned by the model or any off-the-shelf word embedding (e.g. pre-trained word2vec).
We then resize and normalize $\pmb{\hat{\beta}}_t$ in the same way as the strong supervision scenario.

\section{Attention Correctness: Evaluation Metric}
\label{sec:metric}

At each time step in the implicit attention model, the LSTM not only predicts the next word $y_t$ but also generates an attention map $\pmb{\alpha_t} \in \mathbb{R}^L$ across all locations. 
However, the attention module is merely an intermediate step, while the error is only backpropagated from the word-likelihood loss in Equation~\ref{eqn:loss-cap}. 
This opens the question of whether this implicitly-learned attention module is indeed effective.

Therefore in this section we introduce the concept of \textit{attention correctness}, an evaluation metric that quantitatively analyzes the quality of the attention maps generated by the attention-based model. 

\subsection{Definition}
\label{sec:metric-def}

\begin{wrapfigure}{r}{0.19\textwidth}
\vspace{-12mm}
\centering
\includegraphics[width=0.19\textwidth]{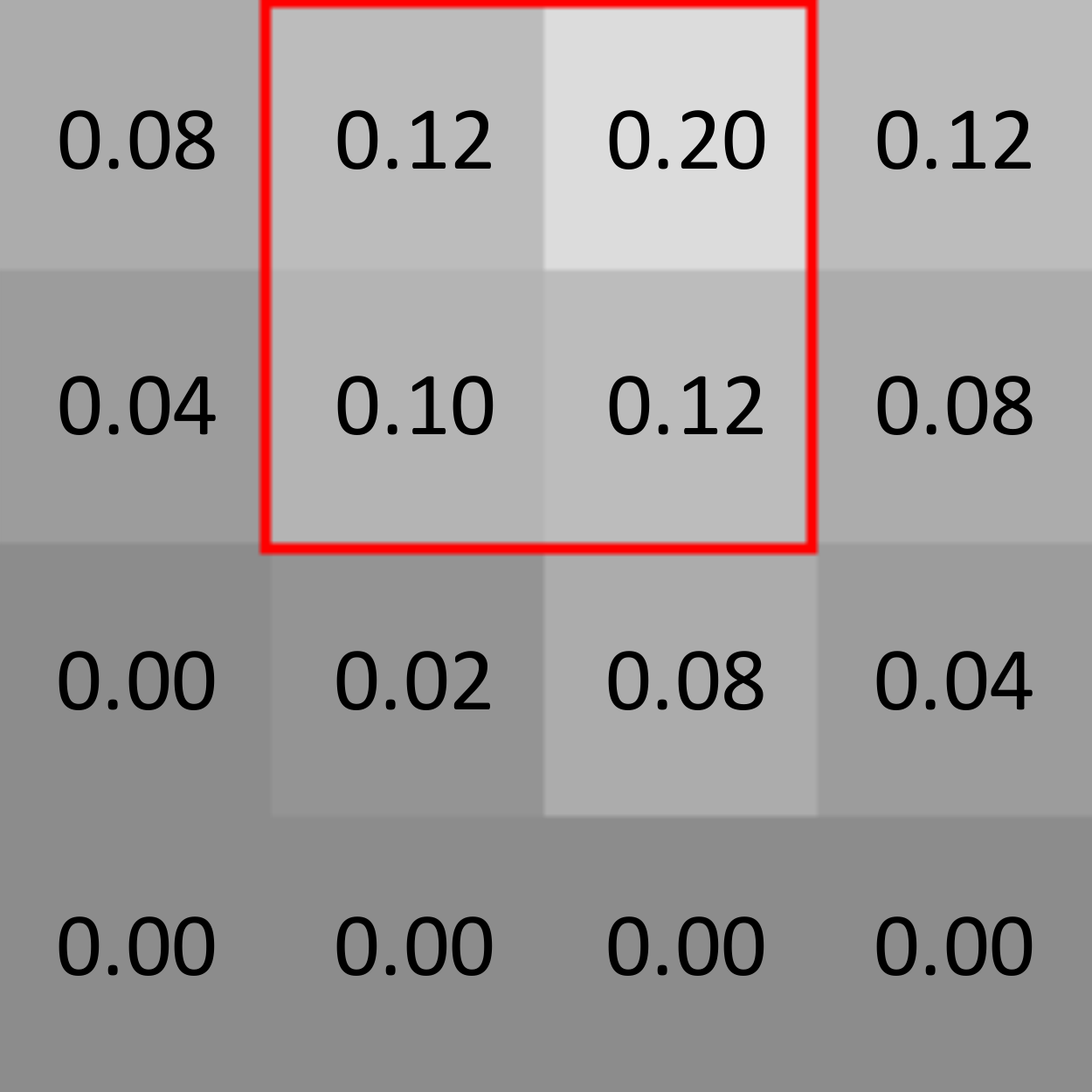}
\caption{Attention correctness is the sum of the weights within ground truth region (red bounding box), in this illustration 0.12 + 0.20 + 0.10 + 0.12 = 0.54.}
\vspace{-4mm}
\label{fig:att-correct}
\end{wrapfigure}

For a word $y_t$ with generated attention map $\pmb{\alpha}_t$, let $R_t$ be the ground truth attention region, then we define the word attention correctness by
\begin{equation}
AC(y_t) = \sum_{\hat{i} \in R_t} \hat{\alpha}_{t\hat{i}}
\end{equation}
which is a score between 0 and 1. Intuitively, this value captures the sum of the attention score that falls within human annotation (see Figure~\ref{fig:att-correct} for illustration). $\pmb{\hat{\alpha}}_t = \{ \hat{\alpha}_{t\hat{i}} \}_{\hat{i} = 1, \hdots, \hat{L}}$ is the resized and normalized $ \pmb{\alpha}_t$ in order to ensure size consistency.

In some cases a phrase $\{y_t, \hdots, y_{t+l} \}$ refers to the same entity, therefore the individual words share the same attention region $R_t$.
We define the phrase attention correctness as the maximum of the individual scores\footnote{In the experiments, we found that changing the definition from maximum to average does not affect our main conclusion.}.
\begin{equation}
AC(\{y_t, \hdots, y_{t+l} \}) = \max (AC(y_t), \hdots, AC(y_{t+l}))
\end{equation}
The intuition is that the phrase may contain some less interesting words whose attention map is ambiguous, and the attention maps of these words can be ignored by the max operation.
For example, when evaluating the phrase ``a group of people'', we are more interested in the attention correctness for ``people'' rather than ``of''. 

% Obviously there are other measures for attention correctness.
% We show in the experimental section that our metric statistically correlates well (spearsman ranking correlation $>$ 0.96) with other metrics, while being the most intuitive.

We discuss next how to find ground truth attention regions during testing, in order to apply this evaluation metric.

\subsection{Ground Truth Attention Region During Testing}
\label{sec:metric-match}

In order to compute attention correctness, we need the correspondence between regions in the image and phrases in the caption. 
However, in the testing stage, the generated caption is often different from the ground truth captions. 
This makes evaluation difficult, because we only have corresponding image regions for the phrases in the ground truth caption, but not \textit{any} phrase.
To this end, we propose two strategies.

\noindent
\textbf{Ground Truth Caption}
One option is to enforce the model to output the ground truth sentence by resetting the input to the ground truth word at each time step. 
This procedure to some extent allows us to ``decorrelate'' the attention module from the captioning component, and diagnose if the learned attention module is meaningful. 
Since the generated caption exactly matches the ground truth, we compute attention correctness for all noun phrases in the test set.

\noindent
\textbf{Generated Caption}
Another option is to align the entities in the generated caption to those in the ground truth caption. 
For each image, we first extract the noun phrases of the generated caption using a POS tagger (e.g. Stanford Parser \cite{manning2014stanford}), and see if there exists a word-by-word match in the set of noun phrases in the ground truth captions.
For example, if the generated caption is ``A dog jumping over a hurdle'' and one of the ground truth captions is ``A cat jumping over a hurdle'', we match the noun phrase ``a hurdle'' appearing in both sentences.
We then calculate the attention correctness for the matched phrases only.

\section{Experiments} 
\label{sec:exp}

\subsection{Implementation Details}

\begin{figure}[t!]
\centering
\captionsetup[subfigure]{labelformat=empty, font=small}
\begin{subfigure}{.23\textwidth}
\captionsetup{width=0.95\textwidth}
\centering
\includegraphics[width=0.32\textwidth]{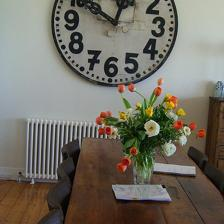}
\includegraphics[width=0.32\textwidth]{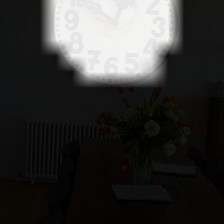}
\includegraphics[width=0.32\textwidth]{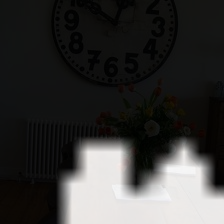}
\caption{\textit{The huge \underline{clock} on the wall is near a wooden \underline{table}.}}
\end{subfigure}
\begin{subfigure}{.23\textwidth}
\captionsetup{width=0.95\textwidth}
\centering
\includegraphics[width=0.32\textwidth]{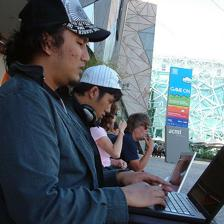}
\includegraphics[width=0.32\textwidth]{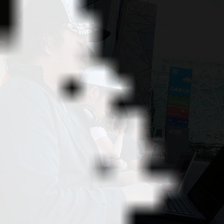}
\includegraphics[width=0.32\textwidth]{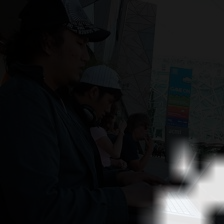}
\caption{\textit{A \underline{man} is on his \underline{laptop} while people looking on. }}
\end{subfigure}
\begin{subfigure}{.23\textwidth}
\captionsetup{width=0.95\textwidth}
\centering
\includegraphics[width=0.32\textwidth]{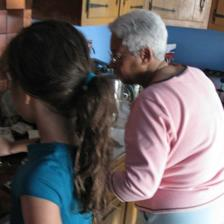}
\includegraphics[width=0.32\textwidth]{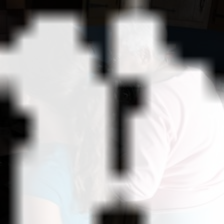}
\includegraphics[width=0.32\textwidth]{figs/coco/girl1}
\caption{\textit{A young \underline{girl} and a \underline{woman} preparing food in a kitchen.}}
\end{subfigure}
\begin{subfigure}{.23\textwidth}
\captionsetup{width=0.95\textwidth}
\centering
\includegraphics[width=0.32\textwidth]{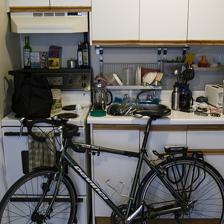}
\includegraphics[width=0.32\textwidth]{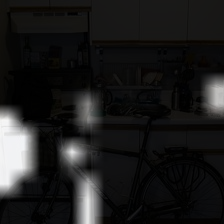}
\includegraphics[width=0.32\textwidth]{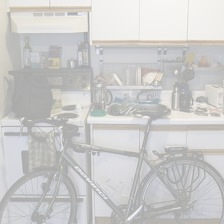}
\caption{\textit{A bicycle parked in a \underline{kitchen} by the stove.}}
\end{subfigure}
\caption{Ground truth attention maps generated for COCO. The first two examples show successful cases. The third example is a failed case where the proposed method aligns both ``girl'' and ``woman'' to the ``person'' category. The fourth example shows the necessity of using the scene category list. If we do not distinguish between object and scene (middle), the algorithm proposes to align the word ``kitchen'' with objects like ``spoon'' and ``oven''. We propose to use uniform attention (right) in these cases.}
\label{fig:coco-gt}
\end{figure}

\textbf{Implicit/Supervised Attention Models} 
All implementation details strictly follow \cite{xu2015show}. 
We resize the image such that the shorter side has 256 pixels, and then center crop the $224\times 224$ image, before extracting the conv5\textunderscore4 feature of the 19 layer version of VGG net \cite{simonyan2014very} pretrained on ImageNet \cite{deng2009imagenet}.
The model is trained using stochastic gradient descent with the Adam algorithm \cite{kingma2014adam}. 
Dropout \cite{srivastava2014dropout} is used as regularization. 
We use the hyperparameters provided in the publicly available code\footnote{https://github.com/kelvinxu/arctic-captions}. 
We set the number of LSTM units to 1300 for Flickr30k and 1800 for COCO.

\noindent
\textbf{Ground Truth Attention for Strong Supervision Model} 
We experiment with our strong supervision model on the Flickr30k dataset \cite{young2014image}. 
The Flickr30k Entities dataset \cite{plummer2015flickr30k} is used for generating the ground truth attention maps. 
For each entity (noun phrase) in the caption, the Flickr30k Entities dataset provides the corresponding bounding box of the entity in the image. 
Therefore ideally, the model should ``attend to'' the marked region when predicting the associated words. 
We evaluate on noun phrases only, because for other types of words (e.g. determiner, preposition) the attention might be ambiguous and meaningless. 

\noindent
\textbf{Ground Truth Attention for Weak Supervision Model} 
The MS COCO dataset \cite{lin2014microsoft} contains instance segmentation masks of 80 classes in addition to the captions, which makes it suitable for our model with weak supervision.
We only construct $\pmb{\beta}_t$ for the nouns in the captions, which are extracted using the Stanford Parser \cite{manning2014stanford}.
The similarity function in Equation~\ref{eqn:sim} is chosen to be the cosine distance between word vectors \cite{mikolov2013distributed} pretrained on GoogleNews\footnote{https://code.google.com/archive/p/word2vec/}, and we set an empirical threshold of 1/3 (i.e. only keep those with cosine distance greater than the threshold). 

The $\pmb{\beta}_t$ generated in this way still contains obvious errors, primarily because word2vec cannot distinguish well between objects and scenes. 
For example, the similarity between the word ``kitchen'' and the object class ``spoon'' is above threshold. 
But when generating a scene word like ``kitchen'', the model should be attending to the whole image instead of focusing on a small object like ``spoon''. 

To address this problem, we refer to the supplement of \cite{lin2014microsoft}, which provides a scene category list containing key words of scenes used when collecting the dataset. Whenever some word in this scene category list appears in the caption, we set $\pmb{\beta}_t$ to be uniform, i.e. equal attention across image. This greatly improves the quality of $\pmb{\beta}_t$ in some cases (see illustration in Figure~\ref{fig:coco-gt}). 

\noindent
\textbf{Comparison of Metric Designs}
To show the legitimacy of our attention correctness metric, we compute the spearsman correlation of our design and three other metrics: negative L1 distance, negative L2 distance, and KL divergence between $\pmb{\hat{\beta}}_t$ and $\pmb{\hat{\alpha}}_t$. 
On the Flickr30k test set with implicit attention and ground truth caption, the spearsman correlations between any two are all above 0.96 (see supplementary material), suggesting that all these measurements are similar.
Therefore our metric statistically correlates well with other metrics, while being the most intuitive.

\subsection{Evaluation of Attention Correctness}

In this subsection, we quantitatively evaluate the attention correctness of both the implicit and the supervised attention model. 
All experiments are conducted on the 1000 test images of Flickr30k. 
We compare the result with a uniform baseline, which attends equally across the whole image. 
Therefore the baseline score is simply the size of the bounding box over the size of the whole image.
The results are summarized in Table~\ref{tab:attn-improve}. 

\begin{table}[t]
\caption{Attention correctness and baseline on Flickr30k test set. Both the implicit and the (strongly) supervised models outperform the baseline. The supervised model performs better than the implicit model in both settings.}
\label{tab:attn-improve}
\centering
\small{
\begin{tabular}{c c c c c}
\bf{Caption} & \bf{Model} & \bf{Baseline} & \bf{Correctness} \\
\hline \\[-8pt]
\multirow{2}{*}{Ground Truth} & Implicit & 0.3214 & 0.3836 \\
& Supervised & 0.3214 & \bf{0.4329} \\
\hline \\[-8pt]
\multirow{2}{*}{Generated} & Implicit & 0.3995 & 0.5202 \\
& Supervised & 0.3968 & \bf{0.5787} \\
\hline \\[-8pt]
\end{tabular}
}
\end{table}

\begin{table}[t]
\caption{Attention correctness and baseline on the Flickr30k test set (generated caption, same matches for implicit and supervised) with respect to bounding box size. The improvement is greatest for small objects.}
\label{tab:size-split}
\centering
\small{
\begin{tabular}{c c c c c}
\bf{BBox Size} & \bf{Model} & \bf{Baseline} & \bf{Correctness}\\
\hline \\[-8pt]
\multirow{2}{*}{Small} & Implicit & 0.1196 & 0.2484 \\
& Supervised & 0.1196 & \bf{0.3682} \\
\hline \\[-8pt]
\multirow{2}{*}{Medium} & Implicit & 0.3731 & 0.5371 \\
& Supervised & 0.3731 & \bf{0.6117} \\
\hline \\[-8pt]
\multirow{2}{*}{Large} & Implicit & 0.7358 & 0.8117 \\
& Supervised & 0.7358 & \bf{0.8255} \\
\hline \\[-8pt]
\end{tabular}
}
\end{table}

\begin{figure}[t]
\centering
\captionsetup[subfigure]{font=small}
\begin{subfigure}{.23\textwidth}
\centering
\includegraphics[width = \textwidth]{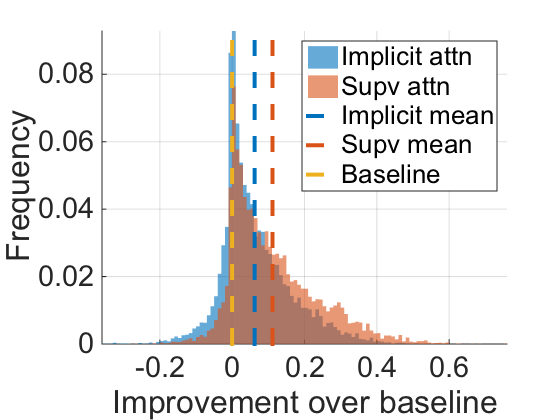}
\caption{Ground truth caption result}
\end{subfigure}
\begin{subfigure}{.23\textwidth}
\centering
\includegraphics[width = \textwidth]{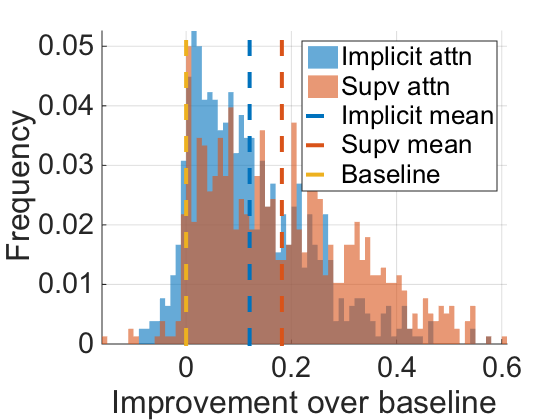}
\caption{Generated caption result}
\end{subfigure}
\caption{Histograms of attention correctness for the implicit model and the supervised model on the Flickr30k test set. The more to the right the better.}
\label{fig:attn-improve}
\end{figure}

\noindent
\textbf{Ground Truth Caption Result}
In this setting, both the implicit and supervised models are forced to produce exactly the same captions, resulting in 14566 noun phrase matches. 
We discard those with no attention region or full image attention (as the match score will be 1 regardless of the attention map). 
For each of the remaining matches, we resize the original attention map from $14\times 14$ to $224\times 224$ and perform normalization before we compute the attention correctness for this noun phrase.

Both models are evaluated in Figure~\ref{fig:attn-improve}a. 
The horizontal axis is the improvement over baseline, therefore a better attention module should result in a distribution further to the right. 
On average, both models perform better than the baseline. 
Specifically, the average gain over uniform attention baseline is 6.22\% for the implicit attention model \cite{xu2015show}, and 11.14\% for the supervised version. 
Visually, the distribution of the supervised model is further to the right. 
This indicates that although the implicit model has captured some aspects of attention, the model learned with strong supervision has a better attention module.

\begin{figure}[t!]
\centering
\captionsetup[subfigure]{labelformat=empty, font=small}
\begin{subfigure}{.23\textwidth}
\centering
\captionsetup{width=0.95\textwidth}
\includegraphics[width=0.32\textwidth]{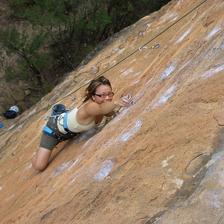}
\includegraphics[width=0.32\textwidth]{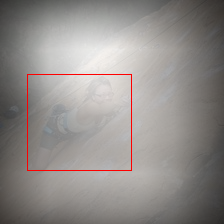}
\includegraphics[width=0.32\textwidth]{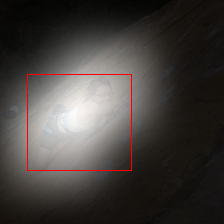}
\caption{\textit{\underline{Girl} rock climbing on the rock wall.}}
\end{subfigure}
\begin{subfigure}{.23\textwidth}
\centering
\captionsetup{width=0.95\textwidth}
\includegraphics[width=0.32\textwidth]{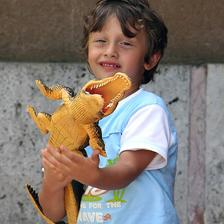}
\includegraphics[width=0.32\textwidth]{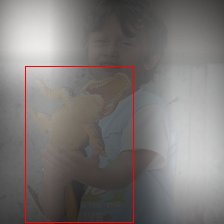}
\includegraphics[width=0.32\textwidth]{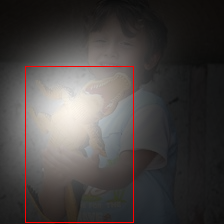}
\caption{\textit{A young smiling child hold his toy \underline{alligator} up to the camera.}}
\end{subfigure}
\begin{subfigure}{.23\textwidth}
\centering
\captionsetup{width=0.95\textwidth}
\includegraphics[width=0.32\textwidth]{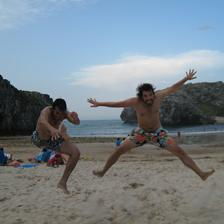}
\includegraphics[width=0.32\textwidth]{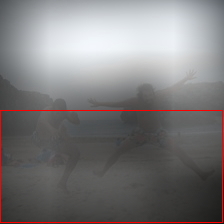}
\includegraphics[width=0.32\textwidth]{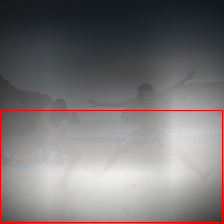}
\caption{\textit{Two male friends in swimming trunks jump on the \underline{beach} while people in the background lay in the sand.}}
\end{subfigure}
\begin{subfigure}{.23\textwidth}
\centering
\captionsetup{width=0.95\textwidth}
\includegraphics[width=0.32\textwidth]{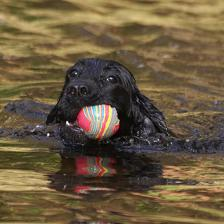}
\includegraphics[width=0.32\textwidth]{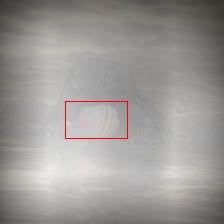}
\includegraphics[width=0.32\textwidth]{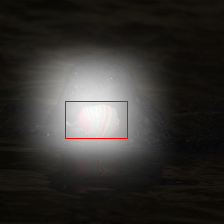}
\caption{\textit{A black dog swims in water with a colorful ball in his \underline{mouth.}}\\\hspace{\textwidth}}
\end{subfigure}
\caption{Attention correctness using ground truth captions. From left to right: original image, implicit attention, supervised attention. The red box marks correct attention region (from Flickr30k Entities). In general the attention maps generated by our supervised model have higher quality.}
\label{fig:gtcap}
\end{figure}

In Figure~\ref{fig:gtcap} we show some examples where the supervised model correctly recovers the spatial location of the underlined entity, while the implicit model attends to the wrong region.

\noindent
\textbf{Generated Caption Result}
In this experiment, word-by-word match is able to align 909 noun phrases for the implicit model and 901 for the supervised version. 
Since this strategy is rather conservative, these alignments are correct and reliable, as verified by a manual check. 
Similarly, we discard those with no attention region or full image attention, and perform resize and normalization before we compute the correctness score.

The results are shown in Figure~\ref{fig:attn-improve}b. 
In general the conclusion is the same: the supervised attention model produces attention maps that are more consistent with human judgment. 
The average improvement over the uniform baseline is 12.07\% for the implicit model and 18.19\% for the supervised model, which is a 50\% relative gain. 

In order to diagnose the relationship between object size and attention correctness, we further split the test set equally with small, medium, and large ground truth bounding box, and report the baseline and attention correctness individually. 
We can see from Table~\ref{tab:size-split} that the improvement of our supervised model over the implicit model is greatest for small objects, and pinpointing small objects is stronger evidence of image understanding than large objects. 

\begin{figure}[t]
\captionsetup[subfigure]{labelformat=empty, font=small}
\begin{subfigure}{.09\textwidth}
\centering
\vspace{-0.2cm}
\caption{Image}
\vspace{-0.2cm}
\includegraphics[width=0.95\textwidth]{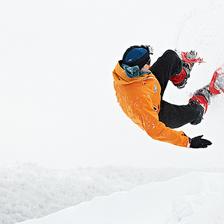}
\caption{\\\hspace{\textwidth}\\\hspace{\textwidth}}
\end{subfigure}
\begin{subfigure}{.185\textwidth}
\centering
\captionsetup{width=0.95\textwidth}
\vspace{-0.2cm}
\caption{Implicit Attention}
\vspace{-0.2cm}
\includegraphics[width=0.46\textwidth]{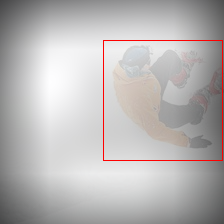}
\includegraphics[width=0.46\textwidth]{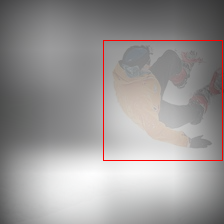}
\caption{\textit{\underline{A man} in a red jacket and blue pants is snowboarding.}}
\end{subfigure}
\begin{subfigure}{.185\textwidth}
\centering
\captionsetup{width=0.95\textwidth}
\vspace{-0.2cm}
\caption{Supervised Attention}
\vspace{-0.2cm}
\includegraphics[width=0.46\textwidth]{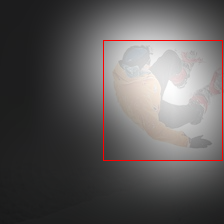}
\includegraphics[width=0.46\textwidth]{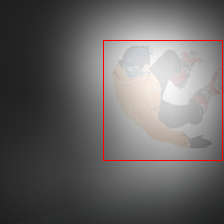}
\caption{\textit{\underline{A man} in a red jumpsuit and a black hat is snowboarding.}}
\end{subfigure}

\begin{subfigure}{.09\textwidth}
\centering
\includegraphics[width=0.95\textwidth]{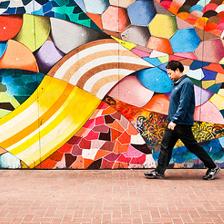}
\caption{\\\hspace{\textwidth}\\\hspace{\textwidth}}
\end{subfigure}
\begin{subfigure}{.185\textwidth}
\centering
\captionsetup{width=0.95\textwidth}
\includegraphics[width=0.46\textwidth]{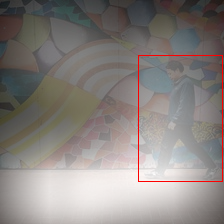}
\includegraphics[width=0.46\textwidth]{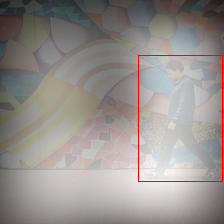}
\caption{\textit{\underline{A man} in a blue shirt and blue pants is sitting on a wall.}}
\end{subfigure}
\begin{subfigure}{.185\textwidth}
\centering
\captionsetup{width=0.95\textwidth}
\includegraphics[width=0.46\textwidth]{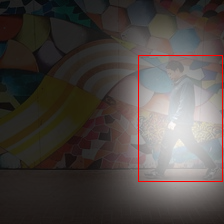}
\includegraphics[width=0.46\textwidth]{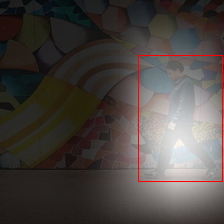}
\caption{\textit{\underline{A man} in a blue shirt and blue pants is skateboarding on a ramp.}}
\end{subfigure}

\begin{subfigure}{.09\textwidth}
\centering
\includegraphics[width=0.95\textwidth]{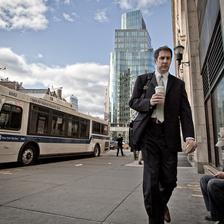}
\caption{\\\hspace{\textwidth}}
\end{subfigure}
\begin{subfigure}{.185\textwidth}
\centering
\captionsetup{width=0.95\textwidth}
\includegraphics[width=0.46\textwidth]{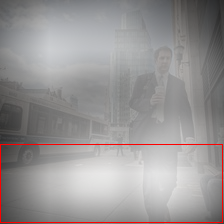}
\includegraphics[width=0.46\textwidth]{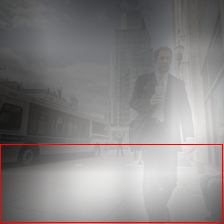}
\caption{\textit{A man and a woman are walking down \underline{the street}.}}
\end{subfigure}
\begin{subfigure}{.185\textwidth}
\centering
\captionsetup{width=0.95\textwidth}
\includegraphics[width=0.46\textwidth]{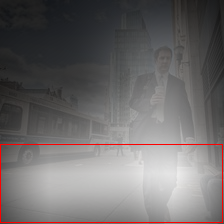}
\includegraphics[width=0.46\textwidth]{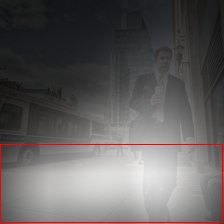}
\caption{\textit{A man and a woman are walking down \underline{the street}.}}
\end{subfigure}

\caption{Attention correctness using generated captions. The red box marks correct attention region (from Flickr30k Entities).  We show two attention maps for the two words in a phrase. In general the attention maps generated by our supervised model have higher quality.}
\label{fig:gcap}
\end{figure}

In Figure~\ref{fig:gcap} we provide some qualitative results. 
These examples show that for the same entity, the supervised model produces more human-like attention than the implicit model. 
More visualization are in the supplementary material.

\subsection{Evaluation of Captioning Performance}

\begin{table}[t]
\caption{Comparison of image captioning performance. * indicates our implementation. Caption quality consistently increases with supervision, whether it is strong or weak.}
\label{tab:cap-improve}
\centering
\small{
\begin{tabular}{c c c c c}
\bf{Dataset} & \bf{Model} & \bf{BLEU-3} & \bf{BLEU-4} & \bf{METEOR} \\
\hline \\[-8pt]
\multirow{3}{*}{Flickr30k} & Implicit & 28.8 & 19.1 & 18.49 \\
% \cline{2-7} \\[-8pt]
&Implicit* & 29.2 & 20.1 & 19.10 \\
% \cline{2-7} \\[-8pt]
&Strong Sup & \bf{30.2} & \bf{21.0} & \bf{19.21} \\
\hline \\[-8pt]
\multirow{3}{*}{COCO} & Implicit &  34.4 & 24.3 & 23.90 \\
% \cline{2-7} \\[-8pt]
&Implicit*  & 36.4 & 26.9 & 24.46 \\
% \cline{2-7} \\[-8pt]
&Weak Sup & \bf{37.2} & \bf{27.6} & \bf{24.78} \\
\hline \\[-8pt]
\end{tabular}
}
\end{table}

\begin{table}[t]
\caption{Captioning scores on the Flickr30k test set for different attention correctness levels in the generated caption, implicit attention experiment. Higher attention correctness results in better captioning performance.}
\label{tab:cap-split}
\centering
\small{
\begin{tabular}{c c c c}
\bf{Correctness} & \bf{BLEU-3} & \bf{BLEU-4} & \bf{METEOR} \\
\hline \\[-8pt]
High & 38.0 & 28.1 & 23.01 \\
Middle & 36.5 & 26.1 & 21.94 \\
Low & 35.8 & 25.4 & 21.14 \\
\hline \\[-8pt]
\end{tabular}
}
\end{table}

We have shown that supervised attention models achieve higher attention correctness than implicit attention models. 
Although this is meaningful in tasks such as region grounding, in many tasks attention only serves as an intermediate step.
We may be more interested in whether supervised attention model also has better captioning performance, which is the end goal. 
The intuition is that a meaningful dynamic weighting of the input vectors will allow later components to decode information more easily.
In this subsection we give experimental support.

We report BLEU \cite{papineni2002bleu} and METEOR \cite{banerjee2005meteor} scores to allow comparison with \cite{xu2015show}. 
In Table~\ref{tab:cap-improve} we show both the scores reported in \cite{xu2015show} and our implementation. 
Note that our implementation of \cite{xu2015show} gives slightly improved result over what they reported.
We observe that BLEU and METEOR scores consistently increase after we introduce supervised attention for both Flickr30k and COCO.
Specifically in terms of BLEU-4, we observe a significant increase of 0.9 and 0.7 percent respectively.

To show the positive correlation between attention correctness and caption quality, we further split the Flickr30k test set (excluding those with zero alignment) equally into three sets with high, middle, and low attention correctness. 
The BLEU-4 scores are 28.1, 26.1, 25.4, and METEOR are 23.01, 21.94, 21.14 respectively (see Table~\ref{tab:cap-split}). 
This indicates that higher attention correctness means better captioning performance.

\section{Discussion}

In this work we make a first attempt to give a quantitative answer to the question: to what extent are attention maps consistent with human perceptions?
We first define attention correctness in terms of consistency with human annotation at both the word level and phrase level. 
In the context of image captioning, we evaluated the state-of-the-art models with implicitly trained attention modules. 
The quantitative results suggest that although the implicit models outperform the uniform attention baseline, they still have room for improvement.

We then show that by introducing supervision of attention map, we can improve both the image captioning performance and attention map quality.
In fact, we observe a positive correlation between attention correctness and captioning quality.
Even when the ground truth attention is unavailable, we are still able to utilize the segmentation masks with object category as a weak supervision to the attention maps, and significantly boost captioning performance.

We believe closing the gap between machine attention and human perception is necessary, and expect to see similar efforts in related fields.

\section{Acknowledgments}

We gratefully acknowledge support from NSF STC award CCF-1231216  and the Army Research Office ARO 62250-CS.
FS is partially supported by NSF IIS-1065243, 1451412, 1513966,  and CCF-1139148.
We also thank Tianze Shi for helpful suggestions in the early stage of this work.

{\bibliographystyle{aaai}
\small{
\bibliography{attn_eval_aaai_final}
}

\end{document}